\documentclass[10pt]{article}

\usepackage{amsmath}
\usepackage{amssymb}

\usepackage{graphicx}

\usepackage{cite}

\usepackage{color} 

\usepackage{booktabs}

\usepackage{multirow} 

\usepackage{subfig}
\usepackage{tabularx}

\PassOptionsToPackage{normalem}{ulem}
\usepackage[normalem]{ulem}

\usepackage{array}

\usepackage{float}
\restylefloat{table}

\date{}

\newcommand{\tabref}[1]{Table~\ref{#1}}

\begin{document}


\title{Extraction of evidence tables from abstracts of randomized clinical trials using a maximum entropy classifier and global constraints}

\author{Antonio Trenta, 
Anthony Hunter\footnote{Corresponding author email: anthony.hunter@ucl.ac.uk}, 
and Sebastian Riedel,\\
Department of Computer Science,\\ University College London,\\ Gower Street, London WC1E 6BT, UK}

\maketitle


\begin{abstract}
Systematic use of the published results of randomized clinical trials is increasingly important in evidence-based medicine. In order to collate and analyze the results from potentially numerous trials, evidence tables are used to represent trials concerning a set of interventions of interest. An evidence table has columns for the patient group, for each of the interventions being compared, for the criterion for the comparison (e.g. proportion who survived after 5 years from treatment), and for each of the results. Currently, it is a labour-intensive activity to read each published paper and extract the information for each field in an evidence table. There have been some NLP studies investigating how some of the features from papers can be extracted, or at least the relevant sentences identified. However, there is a lack of an NLP system for the systematic extraction of each item of information required for an evidence table. We address this need by a combination of a maximum entropy classifier, and integer linear programming. We use the later to handle constraints on what is an acceptable classification of the features to be extracted.  With experimental results, we demonstrate substantial advantages in using global constraints (such as the features describing the patient group, and the interventions, must occur before the features describing the results of the comparison).
\end{abstract}

\begin{flushleft}
{\bf Keywords:} Information extraction; Machine reading of medical information; Automated analysis of clinical evidence; Mining medical literature; Randomized clinical trials; Automated compilation of Evidence tables; Evidence-based medicine.
\end{flushleft}



\section{Introduction}

The systematic use of evidence is already established in healthcare in the form of evidence-based decision making \cite{Sackett:2000}. Much of this evidence is in the form of randomized clinical trials (RCTs) published as articles in medical journals \cite{Hackshaw09}. Unfortunately, the rapidly increasing number of published RCTs on a subject means that it is challenging for a clinician or biomedical researcher to effectively and efficiently acquire and assimilate that evidence.

\subsection{State of the art}

Finding relevant RCTs involves information retrieval techniques applied to the free text and indexing of the RCTs via information providers such as PubMed, Medline, etc. Each of the retrieved RCTs then needs to be read in order to extract information about the RCT concerning the patient class, the interventions being compared in the trial, the outcomes by which the interventions are being compared (e.g. proportion of patients who live more than 5 years after treatment, proportion of patients who do not have a particular side-effect, etc), and the results of the comparison. The extracted information is then used to populate an evidence table. Currently, it is a labour-intensive activity to read each published paper and extract the information for each field in an evidence table. Although many annotation software tools have been developed \cite{Neves2012}, it is still very expensive and time consuming work that requires domain experts. 

Given the large number of published RCTs, a number of researchers have investigated both knowledge-based and statistical NLP techniques for automating the information extraction process. 
Some of these attempts have concentrated on identifying entire sentences containing required information \cite{Chung2009,Boudin2010,Huang2011,McKnight2003,Hirohata2008,Demner-Fushman2006}.
Demner-Fushman et al. \cite{Demner-Fushman2006} employed an ensemble of knowledge-based and statistical classifiers to identify sentences
containing statements about the study outcomes, and Chung \cite{Chung2009} and Boudin et al. \cite{Boudin2010} 
tried to locate sentences containing a reference to the PIO elements (\textit{Participants,} \textit{Intervention},\textit{
Outcome Measure}). The retrieved sentences may then require human intervention or
additional methods to extract the information to populate an evidence table.

Other attempts at applying NLP techniques have gone beyond the sentence level,
with the aim of retrieving fine-grained information, though mostly they have  concentrated on the identification of the patient group and the intervention and control treatments \cite{Kiritchenko2010,Bruijn2008,Summerscales2011,Hansen2008,Hara2007,Chung2009a,Dawes2007,Xu2007,Demner-Fushman2005}. 
These approaches have resulted in good performances in extracting single information items. 
However, they all rely, to various extents, on rule-based and pattern matching approaches, which, although very powerful, can be error-prone and not robust.
For instance, Demner-Fushman \& Lin \cite{Demner-Fushman2005} used manually crafted rules, and DeBruijn et al. \cite{Bruijn2008}, Kiritchenko et al. \cite{Kiritchenko2010}, and Hara \& Matsumoto \cite{Hara2007} used regular expression matching.
Moreover, dependencies between single information elements are not taken into account. We believe that these dependencies bear precious information that can be exploited to improve the performance, robustness, and scalability, of the information extraction system.

\subsection{Overview of our approach}

The aim of this paper is to investigate the feasibility and performance of a classification-based fine-grained information extraction system to obtain information from RCT abstracts obtained from PubMed.
The elements that we want to extract follow the PICO framework \cite{Richardson}: the patient group (P), the intervention (I) and control (C) arms, the outcome measure description and its measurements in the two arms (O). The PICO elements are fundamental for the practice of Evidence Based Medicine \cite{Sackett:2000}.

For this, we identify the words that represent the syntactic head of the phrase containing the information required. We can then extract individual information items using the parse tree of the sentence. In this work we concentrated on the identification of the syntactic heads only. However, once these are identified the extraction of the full phrase is a relatively trivial task. We leave this part to future work.

The output from our system will be used to fill out evidence tables  that can be used by healthcare professionals \cite{HG08}, or collated in databases that can be more easily queried \cite{Sim2000}, or they can be used by analysis tools (e.g. for evidence aggregation \cite{Hunter2012}). 
We give an illustration of input and output to our system in Table \ref{fig:sys_out_example}.

We considered abstracts rather than full text articles because abstracts are the first section readers look at when evaluating a trial and in many cases the only part of the report that they can access to (the full text may require a journal subscription). Moreover, the CONSORT statement \cite{Consort08} requires authors to report all the PICO elements in the abstract, therefore for the purposes of this work the full text article is not needed.
 
The core of our system is completely based on statistical techniques. A Maximum Entropy classifier is first trained using a standard set of linguistic features (\tabref{tab:feats}). The output of the classifier is processed by an optimisation procedure that finds the optimal location of each of the information elements that complies with a set of knowledge-based constraints (\tabref{tab:constraints}).

In contrast, most of the previous attempts to extract fine-grained information rely somehow on rule-based methods (e.g. pattern matching using regular expressions), usually in the final stage (e.g. \cite{Bruijn2008}).
In our system, the use of rules is restricted to the data preprocessing and to the definition of constraints for the inference phase. Therefore, the approach underlying our system can be considered to be a data-driven.

The sentence classification and filtering stage, common in previous work, is not necessary for our system. One reason for this is that most abstracts nowadays are already structured into four or five sections using a labelling policy (e.g. AIM, PATIENTS, METHODS, RESULTS, CONCLUSIONS) with one or more sentences per section. The majority of journals that publish RCTs now adopt a common policy for this labelling such as advised by CONSORT \cite{Consort08} (this was reflected in the set of abstracts retrieved for this study, in which $163$ out of $176$ abstracts had a paragraph structure (\ref{DataCollection})). 
Since, in our system, the abstract structure is used only as a feature to the classifier, and not for filtering out sentences, unstructured abstracts can still be processed. 

Our system takes into account dependencies between the target information items, through the use of abstract-wide constraints in the inference phase. To the best of our knowledge, this is the first attempt to directly consider dependencies between all the PICO elements in an RCT abstract. When in previous work some dependency was considered, it was done either at the sentence level (i.e. \cite{Chung2009}) or for a limited number of elements (i.e. \cite{Summerscales2011}). 
Finally, our approach is the first approach to use the \emph{inference with classifiers} technique \cite{Roth2002} for extracting information from RCTs.

In our test set the system achieved a $0.88$ precision for the patient group, $0.72$ for the intervention and $0.64$ for the control arms, $0.72$ for the outcome measure description and finally $0.68$ and $0.48$ for the outcome measurements in the two arms. The overall precision of the system is $0.687$.

\section{Methods}

In this section, we explain how we obtained our training and testing data from PubMed, and how we annotated this dataset. The dataset concerned interventions for glaucoma and ocular hypertension.

\subsection{Data Collection}\label{DataCollection}

The dataset used to train and test the system contains a number of RCT abstracts, that we retrieved directly from PubMed. 

Following the CONSORT statement 2010 \cite{Schulz2010}, we focused on the most common randomised controlled trial design: individually randomised, two group, parallel.
Different study designs, as well as trials with multiple intervention arms, are reported in different ways, with more complex linguistic structures, as they need varying amounts of additional information. The processing of these trial reports goes beyond the scope of this initial study, therefore we leave it to future work.

We retrieved the abstracts using three search strategies (Appendix A): (Strategy 1) titles or abstracts containing the word ``Glaucoma'' and that specified that the studies were randomized clinical trials; (Strategy 2) titles containing at least one element of a list of prescription drugs recognized as those used typically in the treatment of glaucoma or ocular hypertension and that specified that the studies were randomized clinical trials; and (Strategy 3) titles containing at least one element of a list of surgery procedures,  identified as those typically used in the treatment of glaucoma or ocular hypertension, and that specified that the studies were randomized clinical trials.
The search queries produced $176$ abstracts. 

An abstract was excluded from the dataset if:

\begin{enumerate}
\item the \textbf{study design} did not match the randomized, double-blinded layout;
\item the study involved three or more \textbf{intervention arms} (three or more compared interventions); 
\item the \textbf{results} were not stated clearly with a numerical value (or other formats such as confidence intervals or ranges) for the two arms.
\end{enumerate}

The third condition comes again from the CONSORT statement, in its version for RCT abstracts, in which it is stated that the results should be reported as a summary of the outcome in each group (e.g. the number of participants with or without the event, or the mean and standard deviation of measurements) \cite{Hopewell2008}. This is an increasingly common practice for trial reporting, although sometimes the study results
are still reported in a less schematic and more discursive way (e.g. ``\textit{Bimatoprost
achieved high percentage IOP (intraocular pressure) reductions from baseline in a significantly
higher proportion of patients}'' from \cite{Martin07}).

The restriction of the analysis to a specific medical domain is a common practice in related work (i.e. \cite{Hara2007}), and it was used to reduce  the vocabulary variety in order to make the interpretation of the experimental results simple and clear.
However, this does not exclude that the system can be trained and tested on different medical domains. 

After the filtering stage the dataset contained $99$ abstracts, of which $96$ abstracts were structured (i.e. each abstract was structured into four or five sections using a labelling policy (e.g. AIM, PATIENTS, METHODS, RESULTS, CONCLUSIONS) with one or more sentences per section (such as the CONSORT policy \cite{Consort08}).\footnote{The number of structured abstracts before filtering was $163$.}
A listing of alternative headings and subheadings is given in Table \ref{tab:Example-of-section}. 

The 99 abstracts in our dataset were annotated by hand. This involved enclosing in tags (e.g. $<$tag$>$ ...$<$/tag$>$ ) the syntactic heads of the phrases containing the target information. The tags are listed in \tabref{tab:Annotation-tags}.
All occurrences of patient group, intervention arm, control arm and main outcome measure evaluations were annotated. The main outcome measure descriptor was annotated within the RESULTS section, in the same sentence as the evaluations when possible. 

We have made the dataset publicly available\footnote{http://www0.cs.ucl.ac.uk/staff/a.hunter/projects/crtdata (accessed 30 June 2014)}.

\subsection{Preprocessing}

Various procedures were used to preprocess the dataset. All the proceedures used in preprocessing are automated (i.e. there is no preprocessing by hand). The abstracts
were first split into sentences, using the NLTK package \cite{Bird2009}.

All the abbreviations were substituted with their corresponding expanded version (e.g. ``IOP'' was expanded to ``Intraocular Pressure'').
This was done by either querying a dictionary of known abbreviations or by guessing the right expansion from the text. For the first case, a dictionary was built from the dataset with the most frequent abbreviations encountered with a mapping to their respective expansions. 
When a word that is a key in the dictionary is found in a sentence, it is substituted for its expanded version. 

To account for possible new abbreviations, or different formats of known ones, a function tries to guess possible abbreviations from the abstract taking advantage of the usual way in which they are defined (e.g. “This Is An Example (TIAE)”). The function finds all capitalised words included in brackets, and takes an appropriate number of words preceding them as the expanded version of that abbreviation. This heuristic is designed to be conservative and it performs well with our data, correctly substituting most of the abbreviations\footnote{However, it is unable to take into account pluralised abbreviations (i.e. IOPs) or more complex ones, containing characters like numbers and punctuation marks}.

Numbers and other mathematical constructions were substituted with
an appropriate canonical form, for instance numbers were mapped to
the tag \_NUM\_, percentages to \_PERC\_ and measurement (i.e. ``33
mm'') to \_MEAS\_. But also more complex constructions were replaced
(e.g. confidence intervals to \_CONFINT\_). 
For this
task regular expressions were used to find the target patterns in the
sentence. Patterns were searched multiple times for each sentence,
and this makes it possible to use more complex designs that can, for
instance, solve correctly some kind of ellipsis (i.e. ``10, 20 and
30 mmHg'' becomes ``\_MEAS\_, \_MEAS\_ and \_MEAS\_'' rather than
``\_NUM\_, \_NUM\_ and \_MEAS\_''). In this stage hyphens and double
spaces were also substituted with a single space.

The normalised sentence was then processed by the GENIA tagger \cite{Tsuruoka2005},
which is a well-known system for part-of-speech tagging, NP chunking
and named entity recognition, and it has been designed specifically for biomedical
text.

All chunks in the sentence were given a semantic class label according to their content. For instance ``\textit{Primary open-angle Glaucoma}''
is categorised as DISEASE-OR-MEDICAL-CONDITION (\tabref{tab:Sematic-classes-used}).
This was done by combining a rule-based pattern-matching approach with a search routine in Freebase\footnote{https://developers.google.com/freebase/ (accessed 10 February 2014)}.

Using regular expressions the chunk is first searched for a set of patterns that clearly identify one of the semantic classes in \tabref{tab:Sematic-classes-used}. For instance, if the chunk contains the tag ``\_GONE\_'' or it ends with the word ``group'' or ``arm'', it is given the semantic class ``ARM''. 

If in this first round no semantic class is assigned, the chunk is searched in Freebase. The search routine is designed to avoid inappropriate results. The chunk is searched first in its entirety. If the search does not produce any appropriate result, one word is taken out from the start of the chunk and the remainder is searched again. 

This continues until the chunk is reduced to its rightmost word. Only results referring to one of the semantic classes in \tabref{tab:Sematic-classes-used} are considered, and only if they appear in the top positions of the retrieved list.
If in the end no semantic class is found, the chunk is assigned to the class “none”.

Finally, since the classifier considers single tokens rather than chunks, the chunk semantic class is assigned to the tokens in the chunk, unless the token is a stopword (i.e. and, or, for, not), in which case the semantic class is changed to ``none".

\subsection{System Overview}

The problem of finding the syntactic heads of our target information items is formalized here as a classification problem, in which single tokens in the abstracts are classified in one of the seven classes as listed in \tabref{tab:Annotation-tags}. 
This produces a labelling for each abstract in the form of a sequence of class labels. 

Although we can think of the token labelling as mostly dependent on
local features (e.g. bag-of-words, semantic classes, ...) there are
also global, abstract-wise factors to consider. For instance, we want
to find one and only one instance of each class in each abstract,
the trial results will never appear in the OBJECTIVE section, etc.
Yet, the problem can be seen as a sequential one. Given an abstract,
we ought to find the most likely sequence of labels or, in other words,
the most likely location of the six target labels, that obey 
a set of rules based on general knowledge of how RCTs are reported.

The core of the system consists of two components: a basic classifier
and an inference procedure 
(our code is  freely available\footnote{https://github.com/antoniotre86/IERCT (accessed 23 February 2014)}). 
We follow the \textit{inference with classifiers}
paradigm \cite{Roth2002}: a widely used framework for NLP tasks (e.g. \cite{Roth2003,Punyakanok2004,Clarke2006})
that tackles the problem of combining several local predictors whose
outcomes are mutually dependent. The dependency between predictions
is due to global constraints, which may be related to syntactic or
structural considerations. They may arise from the nature of the problem
or other task-specific conditions. One alternative solution to this problem 
is to incorporate the constraints into a global probabilistic model, 
but this approach is conceptually
more complex and it does not  always seem to perform better experimentally
\cite{TjongKimSang2003}. 

In the \textit{inference with classifiers} framework, inference is carried
out in two separate steps. First, multiple local classifiers are trained,
without any knowledge of the global constraints. In the second step,
the outputs of the local classifiers are processed by an inference
procedure that finds the optimal global prediction subject to the
constraints.

In this work, the \textit{inference with classifiers} framework
has been instantiated in the following way:

\begin{enumerate}

\item A \textbf{maximum entropy classifier} is trained on the dataset, using a set of local and global features. Tokens are classified in one of the seven classes in \tabref{tab:Annotation-tags}. The classifier outputs a vector of scores for each token which represent the probability of that token being in one of the classes; for each token we have a score for each of the six labels. 

\item The optimal labelling (e.g. the one that maximizes the sum of the scores) is found by solving an\textbf{ integer linear programming} (ILP) problem, subject to a set of abstract-wise constraints. This chooses one label for each token such that the sum of such scores over the abstract is maximised, while making sure that none of the constraints are violated. 

\end{enumerate}

To train the maximum entropy classifier a set of local and global features have
been devised (\tabref{tab:feats}). The features represent characteristics of each token
in the abstracts and may be specific to the single token, or to the
chunk, the sentence and the paragraph in which the token is in  
(for instance, the word itself, the part-of-speech
tags, the semantic class, the position, the paragraph label, and whether the word appears
in the title).
The token specific features include also features of the neighbouring
tokens, in a window of one or two to each side. Sentences are treated
as separate entities: when a token is at the start (or at the end)
of the sentence, the features related to the tokens before (after)
are given the value ``none''. 

Before training the classifier, the training set is filtered in order to remove all the tokens that cannot contain any target information. The CONCLUSIONS section is often a concise re-wording of the previous sections, and so it is not likely to contain any PICO element that is not in the rest of the text. For this reason, when a paragraph of category CONCLUSIONS is present, all the tokens in it are filtered out. Moreover, it is reasonable to think that the syntactic heads targets of our classification problem should be nouns or numbers. 
Tokens that have part-of-speech tag in a restricted list (i.e. not nouns or numbers) are excluded from the analysis\footnote{''\('', ''\)'', '':'', ''CC'', ''DT'', ''EX'', ''FW'', ''IN'', ''LS'', ''PRP\$'', ''WDT'', ''WP'', ''RP'', ''TO'', ''PRP'', ''WRB'', ''PDT'', ''WP\$'', ''MD'', ''JJR'', ''JJS''}.

The maximum entropy classifier is trained using the Scikit-learn \cite{scikit-learn} logistic regression module.
When presented with a new token the trained maximum entropy classifier outputs a vector of estimated probabilities of the token being in one of the seven classes in \tabref{tab:Annotation-tags}. Given a new abstract, then, one such vector is estimated for each token and this, along with a set of appropriate constraints, is the input to the inference process.

In this second stage, integer linear programming (ILP) is used to
find the best labelling, in terms of the sum of log-probabilities,
that satisfies a set of knowledge-based constraints. Given the estimated probabilities returned by the classifier, we want to find the sequence of labels that maximizes the sum of the log-probabilities, and satisfied a set of specified constraints.
These constraints arise from considerations concerning the nature of the discourse in a scientific abstract, the way the abstracts were annotated and other aspects that an appropriate annotation should take into account. For instance, since one word cannot be a syntactic head of multiple target information items at a time, we constrained the system to assign only one label per token. As another example, the system is constrained to look for the outcome measure and the results in the RESULTS section, and only after the patient group and the trial arms are reported. A detailed account of all the constraints used can be found in \tabref{tab:constraints}. We also seek to minimize the distance between the A1 and A2 labels and between the R1 and R2 labels. 

If the abstract is unstructured, in order to apply some of the constraints in \tabref{tab:constraints} (in particular (5) and (6)), in the preprocessing phase the paragraph category is guessed by a heuristic that splits the abstracts in five sections of equal length and assigns each section the labels 'BACKGROUND', 'OBJECTIVE', 'METHODS', 'RESULTS' and 'CONCLUSIONS' respectively.

The Gurobi optimizer \cite{gurobi} is used to solve the ILP problem with the constraints listed above. The output of the optimizer is then converted into a proper labelling, integrating also the tokens excluded in the filtering phase (which are given the null label O). This produces a list of estimated labels, one for each token in the abstract. 


\section{Results and discussion}

\subsection{Evaluation}

The performance of the system (\textbf{full model}) has been tested against two baseline models (vanilla and zero models).

The \textbf{vanilla} model is a rudimentary version of the full one.
It still includes the basic maximum entropy classifier but, in the inference phase, it employs a simpler ILP problem which does not consider most of the constraints introduced in the previous section. 
This model simply assigns the proper label to the token that in the abstract has the highest probability of being in that class (excluding the null class O). In other words, it does not take into account the sequential and positioning constraints of the full model.
The maximum entropy classifier, instead, is identical to that in the full model, with also the same features (token, chunk, sentence, paragraph and conjunction features).

The \textbf{zero} model is essentially a maximum entropy classifier. It does not take into account any dependency, and the tokens are classified as one of the classes in \tabref{tab:Annotation-tags} independently from each other.

The zero and vanilla models are used to see whether, and to what extent, the performances improve when we formally include the considerations described in the previous section in the model, that is, when taking into account the dependencies between the labels. 

The performance measures are reported in terms of \textit{precision}, that is 
\[
p_{\ell}=\frac{TP_{\ell}}{CP_{\ell}}\;\;\;\;\;\ell\in\{P,A1,A2,OC,R1,R2\}
\]
and \textit{recall}:
\[
r_{\ell}=\frac{TP_{\ell}}{AP_{\ell}}\;\;\;\;\;\;\ell\in\{P,A1,A2,OC,R1,R2\}
\]
where $TP_{\ell}$ is the number of True Positives, the number of tokens correctly labelled as $\ell$, $CP_{\ell}$ is the number of tokens labelled as $\ell$ by the system and $AP_{\ell}$ is the number of tokens annotated as $\ell$ in the dataset.

Since the constraints 1 and 2 (\tabref{tab:constraints}) impose one and one only target label for each abstract, the number of tokens labelled by the vanilla and full models is fixed, and often smaller than the number of annotated tokens, by construction. This makes the recall measure artificially small for these two models. 

Moreover, the objective of the system is to extract the correct information items, whose number and nature are fixed and we know them in advance. And even when they are reported multiple times, our final goal is to get the right information, regardless of where it is mentioned in the text. Therefore, the only meaningful evaluation measure for this purpose is the precision; the recall measures are reported here just for completeness.

\subsection{Results}

Experiments were run on the datasets collected and annotated as described before.
The dataset was first split into a development set and a test set. The development set included $74$ abstracts and had been used for feature selection, parameter tuning
and constraint testing. The test set had been used to test the models on unseen material, and included the remaining $25$ abstracts. All the three unstructured abstract were included in the test set. 

The features used in the maximum entropy classifier had been first devised and then tested, discarding those that decreased the performance or did not produce any improvement. 
The feature testing had been carried via 10-fold cross validation on the development set. 

A qualitative analysis of the outputs revealed that the most relevant features are the word-related ones: the current and neighbouring words, the sentence and chunk bag-of-words. This is not surprising, as in such a small dataset the diversity of the lexicon is not large. Moreover, having restricted the analysis to glaucoma and ocular hypertension studies, the range of content-specific terms is limited. For instance, the word ``pressure'' (as in ``intraocular pressure'') is very often annotated as the syntactic head of the outcome measure, because it is indeed the typical outcome measure of glaucoma-related RCTs. Other very informative features are the token semantic class, the word-in-title feature and the conjunction features (e.g. semantic
class and paragraph category, and word-in-title and semantic class).

The experimental results are shown in Table \ref{tab:results-precision} in terms of precision, with confidence intervals in Table \ref{tab:results-precision-ci}, and in \tabref{tab:results-recall} in terms of recall. The tests were run both on the development set, via 10-fold cross validation (CV), and on the test set (HO). The differences in precision between the three models have been tested for significance using the paired-rank test \cite{Wilcoxon1945}.

\paragraph{Performance of the zero model.}
The zero model, which does not take into account
any kind of dependency between labels, was generally outperformed by the other two in terms of precision. The differences are highly significant
considering the labelling on the whole ($P<0.01$). However, although very simple, the zero model achieved high precision values with the identification of the patient group ($0.92$), values that are comparable with those produced by the full model.
These results are not surprising though. The patient group was the most predictable item, as it was always expressed in a standard form. Intuitively, it is very reasonable to think that the local features in the maximum entropy classifier play a crucial role in the patient group identification (the word ``patients'', or similar, is a very indicative clue, as well as the semantic class PATIENTS), whilst the positional and sequential constraints do not make a considerable difference. 

It is also worth mentioning the performance of the zero model with the trial arms (labels A1 and A2), as the precisions were comparable to those of the full and the vanilla models. The reason for this comes from the fact that all occurrences of the information items were annotated, and P, A1 and A2 are reported multiple times in most abstracts. The zero model, which is not constrained to choose only one label per abstract, can classify correctly more than one token within an abstract, therefore achieving a higher precision. 

This does not appear to be the case for the outcome measure (OC) and the results (R1 and R2). This may happen because these items are less predictable than the other three, in that they are not usually reported in a standardised way, or with the use of any particular strong cue word or syntactic structure.

\paragraph{Performance of the vanilla model.}
The only element that differentiates the vanilla from the zero model is the constraint that requires only one label instance per abstract (constraint (2) in \tabref{constraints}). 

Looking at the results we can see that this constraint is beneficial for the identification of the OC, R1 and R2 tokens while the precisions for P are lower than the ones achieved by the Zero model and there is no clear advantage with the identification of A1 and A2, with the Vanilla model having higher precisions in the hold out setting, and lower precisions in the cross validation settings. 

The fact that the vanilla model has a lower precision for P, A1 and A2 is mainly due to the fact that these are usually annotated multiple times in the abstract (because the same information is often repeated), in particular P. Since these labels are quite predictable, and repeated multiple times, it is easier for the zero model to identify them correctly, as it often finds at least one candidate and it can make multiple guesses.

The constraint (2) is helpful with those labels for which the classifier has less predictive power (OC, R1 and R2), due to a lack of consistent cue words or linguistic constructs. In many cases, for instance, the Zero model fails to identify a suitable candidate for such labels in the abstract (for no token is the model confident enough to label it as different from the null class O). The vanilla model does not have this problem because it always chooses the most likely token for each label; it introduces a global consideration of the abstract, unlike the zero model which takes into account only the local features.

\paragraph{Performance of the full model.}
The full model outperformed the vanilla and zero models in most cases. It achieved the highest precisions in the test set for all labels: $0.720$ for A1, $0.640$ for A2 (similar to the vanilla model), $0.720$ for OC, $0.680$ for
R1 and $0.480$ for R2. 
Only for label P were the full and vanilla models outperformed by the zero model (as mentioned before).

Compared to the vanilla model, the full model sees larger increases in performance for the results labels (OC, R1 and R2), while for P, A1 and A2 it achieves the same precisions in the HO setting or is slightly better in the CV setting.
This, again, has to do with the fact that OC, R1 and R2 are less predictable than the other labels, and therefore they tend to benefit more from the positional constraints introduced in the full model. 

Conversely, the comparable performances of the full and vanilla models for P, A1 and A2 confirm that the constraints have a weaker impact on these. 

However, we can conclude that the sequential and positional constraints, as well as the minimisation of the distances A1-A2 and R1-R2, have a positive impact on the performances. This suggests that we can use the dependencies between information items to improve the predictive power of the system.

As an example, Table \ref{fig:con1-a} shows the effect that the sequential constraints introduced in the full model have on the labelling of P, A1 and A2. 
In particular, the constraints involved in this example require A1 to be labelled before A2, and P labelled either before or after A1 and A2.
Table \ref{fig:con1-b} shows a snippet of the same abstract, labelled by the vanilla model and by the full model. 

The vanilla model gets the correct token for P (``\emph{patients}'') and A1 (the second `'\emph{Trabeculoplasty}''), but confuses the first ``\emph{trabeculoplasty}'', which has the label A1, for A2. This is because the model is not aware of the fact that the first trial arm is always annotated before the other arm. 

Moreover, here we can clearly see that the two references to the trial arms are both expressed with a coordinating construction. Therefore, since the most sensible
way to label A1 and A2 is to pick them both from the same coordinating construction, P has to be either to the right or to the left of these two. Constraint 6 undertakes this task.
The full model, taking advantage of the additional
information provided by these two constraints, is then able to label correctly both A1 and A2.

Finally, we tested the system on the three unstructured abstracts separately. The zero model correctly identifies two out of three P labels ($0.66$ precision), all three A2 labels and two out of three A1 labels. For labels OC, R1 and R2 it is right in one case only. 
The vanilla model is correct in two out of three abstracts for all labels.
The full model is able to identify the P label in all abstracts, while for the other labels it is correct in two cases.

\section{Conclusion and future work}

This paper describes the development and evaluation of the first phase of an information extraction system, to mine key characteristics of randomized clinical trials from published abstracts on PubMed. The objective of the system is to support, and ultimately substitute, the manual extraction of trial information; an expensive process that is used for various tasks in the healthcare research and practice

The first step towards this objective, presented in this paper, regards the identification of six target information items (the patient group, the intervention and control arms, the main outcome measure and the trial results in each arm) by locating the correct syntactic head of the phrase containing such items. We followed the \emph{inference with classifiers} approach. First, single words in the abstract are classified as one of the six elements with a maximum entropy classifier.
Then the outputs of this first phase are processed by an inference procedure involving the solution of an integer linear programming problem, subject to a set of manually crafted constraints. This approach has the advantage that it can take into account of dependencies between information items, as well as external knowledge of how RCTs evidence
is reported in published abstracts.

The promising results achieved indicate that machine learning can be used in the identification of key elements in RCT abstracts. Moreover, they tell us that dependencies between trial elements can be successfully exploited to obtain better performances and more consistent outputs. We have restricted the domain to one area of medicine, but restricting the domain is a common practice in most of the related work because it reduces the vocabulary variety, thereby making the results clearer. Furthermore, whilst the trained model is indeed domain specific, the model structure is not, and in principle it can be trained to abstracts from different domains without changing the features or the constraints. 

There are number of directions that our system could be developed.
First, the output from our system is the syntactic head of the information we want to extract, and so to extract the required information, the phrase associated with the syntactic heads needs to be extracted from the parse tree.
Secondly, sentences need to be de-normalized, that is,
substituting back the normalization tags with the actual numbers or words. Then, a straightforward improvement could be the substitution of the chunk categorization routine used in the preprocessing phase with more sophisticated techniques, such as the UMLS MetaMap system\footnote{http://www.nlm.nih.gov/research/umls/implementation\_resources/metamap.html (accessed 10 February 2014)} or an ad-hoc statistical method. 
Furthermore, the system should be tested on a larger and more diverse dataset, perhaps extending the analysis to RCTs in different, and more complicated, fields (e.g. oncology).

A possible extension of this work might be the extraction of additional information (i.e. secondary outcomes, eligibility criteria, etc.) and of results reported in a non-schematic way. This would present many challenges.
First of all, with more items to extract it is less likely to find them all in a single abstract. To deal with this issue it may be useful to relax the constraint 2 (i.e. one label per abstract), allowing the system not to identify one or more items. Secondly, the extraction of more verbose result descriptions might require the extraction of entire sentences, and this would undermine the computable representation of the system outputs.
Thirdly, the use of a probabilistic approach may be investigated for this task, with a probabilistic model representing the dependencies between information items. This approach, although more complex than the one presented here, may lead to very interesting results.
Finally, we would like to investigate the automatization of the selection of the abstracts to determine for instance whether an abstract is an RCT and whether it involves two arms.

In the short term, we believe that our approach to automated information extraction would ease the retrieval of relevant evidence from the results of systematic queries. The system
could be used, for instance, to produce a table with the key trial elements from each abstract. The user could then visualise the trial characteristics, and decide if the study is of interest or not, without having to read the whole abstract. 

Our system could also be useful to systematic reviewers who still rely on manual annotation to populate evidence tables. This standard approach usually requires high intellectual effort from domain experts. Although many annotation software tools have been developed \cite{Neves2012}, it is still a very expensive and time consuming work. Automated extraction systems would be then much useful and reduce significantly time and cost of transferring RCT information into evidence tables, or electronic databases, either by supporting human annotation or (ultimately) substituting it.

Finally, an information extraction system could be useful in the context of automated logical reasoning from clinical evidence \cite{Hunter2012}, where knowledge aggregation methods are used to synthesise clinical evidence to produce evidence based recommendations. In such a context,
our information extraction system could ease the process of retrieving and assimilating evidence from full text reports, now based on the expensive manual annotation.

\section{Acknowledgments}

The authors are grateful to Zi Wei Liu and Matthew Williams for feedback on the clinical aspects of this work.
The authors are also grateful to the anonymous referees for a number of valuable suggestions for improving the paper.


\bibliography{references}

\newpage

\section*{Appendix A}
\begin{flushleft}
Glaucoma: 
\par\end{flushleft}

\begin{flushleft}
\texttt{\footnotesize{(clinical trial''{[}Publication
Type{]}) AND (glaucoma{[}Title/Abstract{]}) AND (randomized OR randomised
OR double-masked{[}Title/Abstract{]}) NOT (''protocol''
OR ''non-randomized''{[}Title/Abstract{]})}}
\par\end{flushleft}{\footnotesize \par}

\begin{flushleft}
Prescription drugs: 
\par\end{flushleft}

\begin{flushleft}
\texttt{\footnotesize{(mitomycin{[}Title{]} OR brimonidine{[}Title{]}
OR brinzolamide{[}Title{]} OR dorzolamide{[}Title{]} OR carteolol{[}Title{]}
OR betaxolol{[}Title{]} OR fluorouracil{[}Title{]} OR latanoprost{[}Title{]}
OR bimatoprost{[}Title{]} OR travoprost{[}Title{]} OR timolol{[}Title{]}
AND (randomized{[}Title{]} OR randomised{[}Title{]}) AND (''glaucoma''{[}MeSH
Terms{]} OR ''glaucoma''{[}All Fields{]})}}
\par\end{flushleft}{\footnotesize \par}

\begin{flushleft}
Surgical interventions: 
\par\end{flushleft}

\begin{flushleft}
\texttt{\footnotesize{(randomized{[}Title{]} OR randomised{[}Title{]})
AND (trabeculectomy{[}Title{]} OR phacoemulsification{[}Title{]} OR
trabeculoplasty{[}Title{]} OR phacotrabeculectomy{[}Title{]})}}
\par\end{flushleft}{\footnotesize \par}

\newpage
\section*{Appendix B}\label{appendixB}

To formalize the ILP problem we introduce a set of decision variable 
$\mbox{X=[\ensuremath{\mathbf{x}_{0}},\ensuremath{\mathbf{x}_{1}},...,\ensuremath{\mathbf{x}_{N}}]}$
where $N$ is the number of candidate tokens in the abstract, 
each $\mathbf{x}_{i}$ is equal to 
$[\ensuremath{x_{i,A1}},\ensuremath{x_{i,A2}},\ensuremath{x_{i,P}},\ensuremath{x_{i,OC}},\ensuremath{x_{i,R1}},\ensuremath{x_{i,R2}}]$
and $x_{i,\ell}$ is equal to one if the token $i$ is assigned the label $\ell$, and zero otherwise. 
The following auxiliary variables are introduced:
\begin{itemize}
\item $z_{\ell}\in[0,N],\;\ell\in\mathcal{L} = \{ P, A1, A2, OC, R1, R2, O \}$, to represent the position of label $\ell$ in the abstract, that is \[ z_{\ell}=\sum_{i=1}^{N}x_{i,\ell}\cdot i\;\;\;\;\;\ell\in\mathcal{L} \]
\item $d_{A}$ and $d_{R}$ to represent the distances A1-A2 and R1-R2 respectively,  which are set to be $d_{A}=z_{A2}-z_{A1}$ and $d_{R}=z_{R2}-z_{R1}$;
\item $w_{\ell}\in[0,15]\cup\{101\},\;\ell\in\{R1,R2\}$, which is related to the word of the tokens labelled as R1 and R2. They take values between $0$ and $15$ when the word is equal to one of the sixteen normalization tags, and $101$ (an arbitrary large number) otherwise.  The actual value of $w_{\ell}$ is a mapping to a list of normalization tag types, in which, for instance \_NUM\_ and \_MEAS\_ have the same type. This is to take into account possible ellipses that have not been resolved in the normalization phase. For instance: \texttt{\footnotesize{The mean changes were \_CONFINT\_ in the timolol and \_CONFINTM\_ in the brimonidine group.}}
\item $q_{\ell}\in[0,9],\;\ell\in\{OC,R1,R2\}$, represent the position of the sentence containing the token labelled as $\ell$. The sentence positions are derived from the ``position in sentence'' feature, presented before.
\item $y_{0},y_{1}\in\{0,1\}$, used in the disjunction constraint 7 (\tabref{tab:constraints}). \item $b_{0},b_{1}\in\{0,1\},$ used in the disjunction constraint 10.
\end{itemize}
Thus, our optimization problem can be formalized as 
\begin{equation} 
\max_{X\in\{0,1\}^{7\times N}}\sum_{i=0}^{N}\sum_{\ell\in\mathcal{L}}x_{i,\ell}\log\hat{p}_{i,\ell}-(\delta_{A}d_{A}+\delta_{R}d_{R})\label{eq:linear-problem} 
\end{equation}
where $\hat{p}_{i,\ell}$ is the estimated probability for token $i$ to be assigned the label $\ell$, which are the outputs of the maximum entropy classifier.  
The parameters $\delta_A$ and $\delta_R$ determine the importance of the distance variables in the problem. In this study, they were set to be equal to $10^{-5}$. 
The above is optimized subject to the linear constraints in \tabref{tab:constraints}.

\newpage



\section*{Tables}


\begin{table}[h]
\begin{center}
\begin{tabular}{| l | p{10cm} |}
\hline
Section & Text \\
\hline
\hline
TITLE & Intraocular pressure lowering effect of 0.0015\% tafluprostas compared to placebo in patients with normal tension glaucoma: randomized, double-blind, multicenter, phase III study.\\
\hline
PURPOSE & To compare the efficacy and safety of 0.0015\% tafluprostophthalmic solution (Tafluprost) with Placebo ophthalmic solution (Placebo) in normal tension glaucoma (NTG).\\
\hline
SETTING & Total of 94 patients enrolled in a randomized, double-blind, parallel-group and multi-center study.\\
\hline
METHOD & $\bf \langle P \rangle$ \underline{Patients} with Normal Tension Glaucoma $\bf  \langle /P \rangle$ were randomly assigned to either $\bf  \langle A1 \rangle$ \underline{Tafluprost} $\bf \langle /A1 \rangle$ or $\bf \langle A2 \rangle$ \underline{Placebo } $\bf \langle /A2 \rangle$. Both ophthalmic solutions were instilled once a day in the morning for 4 weeks.\\
\hline
RESULTS & $\bf \langle OC \rangle$ Mean intraocular pressure (IOP) \underline{changes} from baseline $\bf \langle /OC \rangle$ were $\bf \langle R1 \rangle$ \underline{-4.0 +/-1.7 mmHg} $\bf \langle /R1 \rangle$ in Tafluprostadministered patients and $\bf \langle R2 \rangle$ \underline{-1.4 +/-1.8 mmHg} $\bf \langle /R2 \rangle$ in Placebo administered patients at 4 weeks, with a statistically significant difference (p<0.001).\\
\hline
\end{tabular}
\end{center}
\caption[An example of input and output for our system]{Given the abstract as input \cite{Kuwayama10}, 
our system identifies the syntactic heads for each of the tagged items. 
Each syntactic head is underlined. 
The XML tags are $P$ for patient group, $A1$ for arm 1 intervention, $A2$ for arm 2 intervention, $OC$ for outcome being measured, $R1$ outcome for arm 1, and $R2$ outcome for arm 2.
\label{fig:sys_out_example}}
\end{table}


\begin{table}[h]
\begin{centering}
\begin{tabular}{|l|p{10cm}|}
\hline 
\textbf{\small{Section Class}} & \multirow{1}{10cm}{\textbf{\small{Section Label}}}\tabularnewline
\hline 
\hline 
{\footnotesize{BACKGROUND}} & {\footnotesize{INTRODUCTION, TRIAL REGISTRATION, BACKGROUND, FINANCIAL
DISCLOSURE(S), CLINICAL TRIAL REGISTRATION}}\tabularnewline
\hline 
{\footnotesize{OBJECTIVE}} & {\footnotesize{AIMS, BACKGROUND/AIMS, AIM, PURPOSE, OBJECTIVE, INTRODUCTION
AND PURPOSE}}\tabularnewline
\hline 
{\footnotesize{METHODS}} & {\footnotesize{SETTING, PATIENTS AND METHODS, METHODS, STUDY DESIGN
AND METHODS, RESEARCH DESIGN AND METHODS, STATISTICS, SUBJECTS AND
METHODS, METHOD, PARTICIPANTS, MAIN OUTCOME MEASURES, DESIGN, OUTCOME
MEASUREMENT, INTERVENTIONS, MATERIALS AND METHODS, INTERVENTION}}\tabularnewline
\hline 
{\footnotesize{RESULTS}} & {\footnotesize{RESULTS, FINDINGS, MAIN RESULTS}}\tabularnewline
\hline 
{\footnotesize{CONCLUSIONS}} & {\footnotesize{APPLICATION TO CLINICAL PRACTICE, CONCLUSION, CONCLUSIONS,
DISCUSSION}}\tabularnewline
\hline 
\end{tabular}
\par\end{centering}
\caption{Example of section labels and classes in PubMed abstracts found in
the dataset.}\label{tab:Example-of-section}
\end{table}


\begin{table}[m]
\begin{centering}
\begin{tabular}{|l|l|l|}
\hline 
Tag & Description & Examples\tabularnewline
\hline 
\hline 
P & Patient group & \emph{``patients with uncontrolled intraocular pressure''}\tabularnewline
\hline 
A1 & Intervention arm & \textit{``5-fluorouracil''}\tabularnewline
\hline 
A2 & Control arm & \textit{``mitomycin C''}\tabularnewline
\hline 
OC & Outcome being measured & \textit{``number of eyes achieving target pressures''}\tabularnewline
\hline 
R1 & Outcome of intervention arm 1 & \textit{``53 of 56''}\tabularnewline
\hline 
R2 & Outcome of intervention arm 2 & \textit{``54 of 57''}\tabularnewline
\hline 
O & Null class & [Tokens not belonging to any of the above] \tabularnewline
\hline 
\end{tabular}
\end{centering}
\caption{Annotation tags.}\label{tab:Annotation-tags}
\end{table}


\begin{table}[m]
\begin{centering}
\begin{tabular}{|l|l|l|}
\hline 
{\footnotesize{Class label}} & {\footnotesize{Description}} & {\footnotesize{Examples}}\tabularnewline
\hline 
\hline 
\texttt{\footnotesize{ARM}} & {\footnotesize{Reference to the trial arms}} & \textit{\footnotesize{"the timolol group",
"\_GONE\_"}}\tabularnewline
\hline 
\texttt{\footnotesize{CLINICAL-TRIAL}} & {\footnotesize{Clinical trial}} & \textit{\footnotesize{"randomized controlled trial"}}\tabularnewline
\hline 
\texttt{\footnotesize{DIAGNOSTIC-TEST}} & {\footnotesize{Diagnostic test}} & \textit{\footnotesize{"visual test", "tonometry"}}\tabularnewline
\hline 
\texttt{\footnotesize{DISEASE-OR-MEDICAL-CONDITION}} & {\footnotesize{Disease or similar}} & \textit{\footnotesize{"primary open-angle glaucoma"}}\tabularnewline
\hline 
\texttt{\footnotesize{FREQUENCY}} & {\footnotesize{Frequency}} & \textit{\footnotesize{"once daily"}}\tabularnewline
\hline 
\texttt{\footnotesize{MEDICAL-TREATMENT}} & {\footnotesize{Drug, surgery or similar}} & \textit{\footnotesize{"timolol", "trabeculectomy"}}\tabularnewline
\hline 
\texttt{\footnotesize{OUTCOME-MEASURE}} & {\footnotesize{Measurement description}} & \textit{\footnotesize{"intraocular pressure reduction"}}\tabularnewline
\hline 
\texttt{\footnotesize{PATIENTS}} & {\footnotesize{Subjects of the study}} & \textit{\footnotesize{"patients", "subjects",
"women"}}\tabularnewline
\hline 
\texttt{\footnotesize{PERIOD-OF-TIME}} & {\footnotesize{Time interval}} & \textit{\footnotesize{"\_NUM\_ hours",
"\_POFT\_"}}\tabularnewline
\hline 
\end{tabular}
\par\end{centering}
\caption{Semantic classes used in the chunk categorization.}\label{tab:Sematic-classes-used}
\end{table}


\begin{table}[h]
\begin{centering}
\begin{tabular}{|>{\raggedright}p{2cm}|>{\raggedright}p{4cm}|>{\raggedright}p{6cm}|}
\hline 
Feature & Description & Example\tabularnewline
\hline 
\hline 
{\small{Word}} & {\small{The current word}} & {\small{``268 }}\textbf{\uline{\small{patients}}}{\small{ with
ocular hypertension''}}\tabularnewline
\hline 
{\small{Bigram}} & {\small{The previous and current words}} & {\small{``}}\textbf{\uline{\small{268 patients}}}{\small{ with
ocular hypertension''}}\tabularnewline
\hline 
{\small{POS tags}} & {\small{The previous, current and next Part-of-speech tags}} & {\small{``268{[}}}\textbf{\uline{\small{CD}}}{\small{{]} patients{[}}}\textbf{\uline{\small{NN}}}{\small{{]}
with{[}}}\textbf{\uline{\small{IN}}}{\small{{]} ocular{[}JJ{]}
hypertension{[}NN{]}''}}\tabularnewline
\hline 
{\small{Semantic class}} & {\small{The semantic class of the current word}} & {\small{``268 patients{[}}}\textbf{\uline{\small{PATIENTS}}}{\small{{]}
with \{ocular hypertension\}{[}DISEASE{]}}}\tabularnewline
\hline 
{\small{Word in title}} & {\small{Either TRUE or FALSE if the current word appears in the title}} & {\small{``268{[}F{]} patients{[}}}\textbf{\uline{\small{T}}}{\small{{]}
with{[}T{]} ocular{[}T{]} hypertension{[}T{]}''}}\tabularnewline
\hline 
{\small{Word inside brackets}} & {\small{Either TRUE or FALSE if the current word appears inside brackets}} & {\small{``268{[}F{]} patients{[}}}\textbf{\uline{\small{F}}}{\small{{]}
with{[}F{]} ocular{[}F{]} hypertension{[}F{]}''\linebreak``Intraocular{[}F{]}
pressure{[}F{]} was{[}F{]} significantly{[}F{]} (P\textless{}0.001{[}T{]})
reduced{[}T{]}''}}\tabularnewline
\hline 
{\small{Position in sentence}} & {\small{Relative position of the current word in the sentence}} & {\small{``268{[}0{]} patients{[}}}\textbf{\uline{\small{3}}}{\small{{]}
with{[}5{]} ocular{[}7{]} hypertension{[}9{]}''}}\tabularnewline
\hline 
{\small{Chunk bag of words}} & {\small{Words included in the current chunk}} & {\small{\{''268'', ``patients''\}}}\tabularnewline
\hline 
{\small{Chunk type}} & {\small{Whether the current chunk is a noun phrase, a verb phrase etc.}} & {\small{``\{268 patients\}{[}}}\textbf{\uline{\small{NP}}}{\small{{]}
with{[}PP{]} \{ocular hypertension\}{[}NP{]}''}}\tabularnewline
\hline 
{\small{Sentence position}} & {\small{Position of the current sentence in the paragraph}} & {\small{-}}\tabularnewline
\hline 
{\small{Paragraph label}} & {\small{Title given to the current paragraph by the authors}} & {\small{``PATIENTS AND METHODS''}}\tabularnewline
\hline 
{\small{Paragraph category}} & {\small{The category assigned by PubMed to the current paragraph}} & {\small{``METHODS''}}\tabularnewline
\hline 
{\small{Conjunction features}} & {\small{Features combining some of the above}} & {\small{-}}\tabularnewline
\hline 
\end{tabular}\caption{Features used to train the maximum entropy classifier.}\label{tab:feats}
\par\end{centering}
\end{table}


\begin{table}[h]
\begin{centering}
\begin{tabular}{|>{\raggedright}p{4cm}|>{\raggedright}p{5cm}|>{\raggedright}p{5cm}|}
\hline 
Constraint & Motivation & Formally\tabularnewline
\hline 
\hline 
{\footnotesize{1) Each token can be assigned one label only}} & {\footnotesize{A word cannot be the syntactic head of multiple target
information items at a time}} & {\footnotesize{$\sum_{\ell\in{\cal L}}x_{i,\ell}=1,\;\;\; i=0,...,N$ }}\tabularnewline
\hline 
{\footnotesize{2) For each abstract, one token, and one only, is assigned
a label different from the null class ${\cal O}$}} & {\footnotesize{In our analysis all abstracts have at least one reference
for each information item, we take only one to keep the analysis simple}} & {\footnotesize{$\sum_{i=0}^{N}x_{i,\ell}=1,\;\;\;\forall\ell\in{\cal L}\setminus\{{\cal O}\}$}}\tabularnewline
\hline 
{\footnotesize{3) A1 must appear before A1}} & {\footnotesize{The trial arms are always annotated in the order A1
$\rightarrow$A2}} & {\footnotesize{$z_{A2}-z_{A1}\geq0$}}\tabularnewline
\hline 
{\footnotesize{4) A1, A2 and P must appear before OC, R1 and R2}} & {\footnotesize{The results of the study are always reported after
the problem setting}} & {\footnotesize{$z_{\ell_{1}}-z_{\ell_{0}}\geq0,\;\;\;\ell_{1}\in\{OC,R1,R2\},\ell_{0}\in\{A1,A2,P\}$}}\tabularnewline
\hline 
{\footnotesize{5) R1 and R2 cannot appear outside the RESULTS section}} & {\footnotesize{If the abstract has a structure, R1 and R2 appear always
in the RESULTS section}} & {\footnotesize{$x_{i,\ell}=0,\;\;\; i\in{\cal R},\ell\in\{R1,R2\}$\linebreak ${\cal R}:=\{j:tok_{j}("paragraph")\notin\{RESULTS,none\},j=1,...,N\}$}}\tabularnewline
\hline 
{\footnotesize{6) A1, A2 and P cannot appear in the RESULTS section}} & {\footnotesize{If the abstract has a structure, P, A1 and A2 never
appear in the RESULTS section}} & {\footnotesize{$x_{i,\ell}=0,\;\;\; i\in{\cal M},\ell\in\{A1,A2,P\}$\linebreak${\cal M}:=\{j:tok_{j}("paragraph")=RESULTS,j=1,...,N\}$}}\tabularnewline
\hline 
{\footnotesize{7) P cannot appear between A1 and A2}} & {\footnotesize{The patient group is always reported before or after
the trial arms}} & {\footnotesize{$(z_{A1}-z_{P})y_{0}\geq0$\linebreak$(z_{P}-z_{A2})y_{1}\geq0$\linebreak$y_{0}+y_{1}=1$}}\tabularnewline
\hline 
{\footnotesize{8) Tokens labelled as R1 and R2 must be numerical quantities
(including ranges, confidence intervals etc)}} & {\footnotesize{The results are always represented in numerical terms}} & {\footnotesize{$w_{\ell}\leq100\;\;\;\ell\in\{R1,R2\}$}}\tabularnewline
\hline 
{\footnotesize{9) Tokens labelled as R1 and R2 must be of the same
quantity}} & {\footnotesize{The results must be comparable quantities}} & {\footnotesize{$w_{R1}-w_{R2}=0$}}\tabularnewline
\hline 
{\footnotesize{10) The sentence containing OC must have the same position
in the paragraph as the sentence containing R1 and R2}} & {\footnotesize{The outcome measure is always reported in the same
sentence as the results}} 
& {\footnotesize{$(q_{OC}-q_{R1})b_{0}\geq0$\linebreak$(q_{OC}-q_{R2})b_{1}\geq0$\linebreak $b_{0}+b_{1}=1$}}\tabularnewline
\hline 
\end{tabular}
\par\end{centering}
\caption{Constraints in the ILP problem.}\label{tab:constraints}
\end{table}

\hspace{5cm}

\clearpage

\begin{table}[h]
\begin{centering}
\begin{tabular}{llcccccc}
\toprule 
 & 
 & 
{\footnotesize{P}}
 & 
{\footnotesize{A1}}
 & 
{\footnotesize{A2}}
 & 
{\footnotesize{OC}}
 & 
{\footnotesize{R1}}
 & 
{\footnotesize{R2}}
\tabularnewline
\midrule
\midrule 
\multirow{3}{*}{{\footnotesize{CV}}} & \textbf{\footnotesize{zero}} & {\footnotesize{0.919}} & {\footnotesize{0.797}} & {\footnotesize{0.716}} & {\footnotesize{0.270}} & {\footnotesize{0.270}} & {\footnotesize{0.162}}\tabularnewline
\cmidrule{2-8} 
 & \textbf{\footnotesize{vanilla}} & {\footnotesize{0.905}} & {\footnotesize{0.703}} & {\footnotesize{0.622}} & {\footnotesize{0.635}} & {\footnotesize{0.514}} & {\footnotesize{0.365}}\tabularnewline
\cmidrule{2-8} 
 & \textbf{\footnotesize{full}} & {\footnotesize{0.932}} & {\footnotesize{0.784}} & {\footnotesize{0.649}} & {\footnotesize{0.622}} & {\footnotesize{0.595}} & {\footnotesize{0.486}}\tabularnewline
\midrule 
\multirow{3}{*}{{\footnotesize{HO}}} & \textbf{\footnotesize{zero}} & {\footnotesize{0.920}} & {\footnotesize{0.560}} & {\footnotesize{0.560}} & {\footnotesize{0.200}} & {\footnotesize{0.280}} & {\footnotesize{0.200}}\tabularnewline
\cmidrule{2-8} 
 & \textbf{\footnotesize{vanilla}} & {\footnotesize{0.880}} & {\footnotesize{0.640}} & {\footnotesize{0.640}} & {\footnotesize{0.680}} & {\footnotesize{0.560}} & {\footnotesize{0.440}}\tabularnewline
\cmidrule{2-8} 
 & \textbf{\footnotesize{full}} & {\footnotesize{0.880}} & {\footnotesize{0.720}} & {\footnotesize{0.640}} & {\footnotesize{0.720}} & {\footnotesize{0.680}} & {\footnotesize{0.480}}\tabularnewline
\bottomrule
\end{tabular}
\par\end{centering}
\caption[Precision results for zero, vanilla and full models ]{Precision results for zero, vanilla and full models, computed via cross-validation on the development set (CV), or via hold-out on the test set (HO).\label{tab:results-precision}}
\end{table}


\begin{table}[h]
\begin{centering}
\begin{tabular}{llcccccc}
\toprule 
 & 
 & 
{\footnotesize{P}}
 & 
{\footnotesize{A1}}
 & 
{\footnotesize{A2}}
 & 
{\footnotesize{OC}}
 & 
{\footnotesize{R1}}
 & 
{\footnotesize{R2}}
\tabularnewline
\midrule
\midrule 
 & \textbf{\footnotesize{zero}} & {\footnotesize{0.537 - 0.931}} & {\footnotesize{0.268 - 0.588}} & {\footnotesize{0.261 - 0.55}} & {\footnotesize{0.0642 - 0.373}} & {\footnotesize{-0.0602 - 0.509}} & {\footnotesize{-0.0569 - 0.322}}\tabularnewline
 & \textbf{\footnotesize{vanilla}} & {\footnotesize{0.587 - 0.859}} & {\footnotesize{0.277 - 0.536}} & {\footnotesize{0.219 - 0.503}} & {\footnotesize{0.442 - 0.827}} & {\footnotesize{0.087 - 0.864}} & {\footnotesize{0.216 - 0.585}}\tabularnewline
 & \textbf{\footnotesize{full}} & {\footnotesize{0.597 - 0.873}} & {\footnotesize{0.344 - 0.593}} & {\footnotesize{0.265 - 0.576}} & {\footnotesize{0.418 - 0.942}} & {\footnotesize{0.479 - 0.722}} & {\footnotesize{0.414 - 0.599}}\tabularnewline
\bottomrule
\end{tabular}
\par\end{centering}
\caption[Cross Validation confidence intervals for precision results for zero, vanilla and full models ]{Cross Validation 95\% confidence intervals for the precision results for zero, vanilla and full models.\label{tab:results-precision-ci}}
\end{table}


\begin{table}[h]
\begin{centering}
\begin{tabular}{llcccccc}
\toprule 
 & 
 & 
{\footnotesize{P}}
 & 
{\footnotesize{A1}}
 & 
{\footnotesize{A2}}
 & 
{\footnotesize{OC}}
 & 
{\footnotesize{R1}}
 & 
{\footnotesize{R2}}
\tabularnewline
\midrule
\midrule 
\multirow{3}{*}{{\footnotesize{CV}}} & \textbf{\footnotesize{zero}} & 
{\footnotesize{0.812}} & 
{\footnotesize{0.750}} & 
{\footnotesize{0.400}} & 
{\footnotesize{0.300}} & 
{\footnotesize{0.272}} & 
{\footnotesize{0.363}}\tabularnewline
\cmidrule{2-8} 
 & \textbf{\footnotesize{vanilla}} & 
{\footnotesize{0.625}} & 
{\footnotesize{0.500}} & 
{\footnotesize{0.350}} & 
{\footnotesize{0.500}} & 
{\footnotesize{0.545}} & 
{\footnotesize{0.454}}\tabularnewline
\cmidrule{2-8} 
 & \textbf{\footnotesize{full}} & 
{\footnotesize{0.625}} & 
{\footnotesize{0.550}} & 
{\footnotesize{0.300}} & 
{\footnotesize{0.500}} & 
{\footnotesize{0.545}} & 
{\footnotesize{0.454}}\tabularnewline
\midrule 
\multirow{3}{*}{{\footnotesize{HO}}} & \textbf{\footnotesize{zero}} & 
{\footnotesize{0.657}} & 
{\footnotesize{0.340}} & 
{\footnotesize{0.340}} & 
{\footnotesize{0.200}} & 
{\footnotesize{0.185}} & 
{\footnotesize{0.185}}\tabularnewline
\cmidrule{2-8} 
 & \textbf{\footnotesize{vanilla}} & 
{\footnotesize{0.657}} & 
{\footnotesize{0.340}} & 
{\footnotesize{0.340}} & 
{\footnotesize{0.560}} & 
{\footnotesize{0.407}} & 
{\footnotesize{0.370}}\tabularnewline
\cmidrule{2-8} 
 & \textbf{\footnotesize{full}} & 
{\footnotesize{0.628}} & 
{\footnotesize{0.363}} & 
{\footnotesize{0.363}} & 
{\footnotesize{0.600}} & 
{\footnotesize{0.555}} & 
{\footnotesize{0.444}}\tabularnewline
\bottomrule
\end{tabular}
\par\end{centering}

\caption[Recall results for zero, vanilla and full models]{Recall results for zero, vanilla and full models, computed
via cross-validation on the development set (CV), or via hold-out on the test set (HO).\label{tab:results-recall}}
\end{table}


\clearpage

\begin{table}[h]
\begin{centering}
\subfloat[Vanilla model.\label{fig:con1-a}]{\begin{raggedright}
\framebox{\begin{minipage}[t]{1\columnwidth}%
\texttt{\footnotesize{To compare outcomes of selective laser }}\texttt{\uline{\footnotesize{trabeculoplasty(A1|A2)}}}\texttt{\footnotesize{
with drug therapy(A2|O) for glaucoma }}\texttt{\textbf{\uline{\footnotesize{patients(P|P)}}}}\texttt{\footnotesize{
in a prospective randomized clinical trial NUM NUM patients (NUM eyes)
with open angle glaucoma or ocular hypertension were randomized to
Selective Laser }}\texttt{\textbf{\uline{\footnotesize{Trabeculoplasty(A1|A1)}}}}\texttt{\footnotesize{
or medical therapy(A2|O) Target intraocular pressure {[}...{]}}}
\end{minipage}}
\par\end{raggedright}
}
\par\end{centering}
\begin{centering}
\subfloat[Full model.\label{fig:con1-b}]{%
\framebox{\begin{minipage}[t]{1\columnwidth}%
\texttt{\footnotesize{To compare outcomes of selective laser }}\texttt{\uline{\footnotesize{\textbf{trabeculoplasty(A1|A1)}}}}\texttt{\footnotesize{
with drug }}\texttt{\textbf{\uline{\footnotesize{therapy(A2|A2)}}}}\texttt{\footnotesize{
for glaucoma }}\texttt{\textbf{\uline{\footnotesize{patients(P|P)}}}}\texttt{\footnotesize{
in a prospective randomized clinical trial NUM NUM patients (NUM eyes)
with open angle glaucoma or ocular hypertension were randomized to
Selective Laser Trabeculoplasty(A1|O) or medical therapy(A2|O) Target
intraocular pressure {[}...{]}}}
\end{minipage}}
}
\par\end{centering}
\caption[Illustration of how the constraints in full model improve the labeling]{Illustration of how  the constraints in full model improve the labeling with text from \cite{Katz12}.
The terms in parentheses indicate
the actual label and the label assigned by the system, for instance,
``\texttt{\footnotesize{(A1|A2)}}'' means that the actual label
of the token is A1 and the system labelled it as A2. The tokens that
have been assigned a label different from O are underlined.
\label{fig:How-the-constraints1}}
\end{table}

\end{document}